\documentclass[a4paper]{report}
\usepackage[utf8]{inputenc}
\usepackage[T1]{fontenc}
\usepackage{RJournal}
\usepackage{amsmath,amssymb,array}
\usepackage{booktabs}

\usepackage{longtable}

\makeatletter
 \let\@cite@ofmt\@firstofone
 \def\@biblabel#1{}
 \def\@cite#1#2{{#1\if@tempswa , #2\fi}}
\makeatother
\newlength{\cslhangindent}
\newlength{\csllabelwidth}
 {\begin{list}{}{%
  \setlength{\itemindent}{0pt}
  \setlength{\leftmargin}{0pt}
  \setlength{\parsep}{0pt}
  \ifodd #1
   \setlength{\leftmargin}{\cslhangindent}
   \setlength{\itemindent}{-1\cslhangindent}
  \fi
  \setlength{\itemsep}{#2\baselineskip}}}
 {\end{list}}
\usepackage{calc}

\usepackage{pifont}

\begin{document}

\sectionhead{Contributed research article}
\volume{XX}
\volnumber{YY}
\year{20ZZ}
\month{AAAA}

\begin{article}
\title{xplainfi: Feature Importance and Statistical Inference for Machine Learning in R}

\author{by Lukas Burk, Fiona Katharina Ewald, Giuseppe Casalicchio, Marvin N. Wright, and Bernd Bischl}

\maketitle

\abstract{%
We introduce xplainfi, an R package built on top of the mlr3 ecosystem for global, loss-based feature importance methods for machine learning models. Various feature importance methods exist in R, but significant gaps remain, particularly regarding conditional importance methods and associated statistical inference procedures. The package implements permutation feature importance, conditional feature importance, relative feature importance, leave-one-covariate-out, and generalizations thereof, and both marginal and conditional Shapley additive global importance methods. It provides a modular conditional sampling architecture based on Gaussian distributions, adversarial random forests, conditional inference trees, and knockoff-based samplers, which enable conditional importance analysis for continuous and mixed data. Statistical inference is available through multiple approaches, including variance-corrected confidence intervals and the conditional predictive impact framework. We demonstrate that xplainfi produces importance scores consistent with existing implementations across multiple simulation settings and learner types, while offering competitive runtime performance. The package is available on CRAN and provides researchers and practitioners with a comprehensive toolkit for feature importance analysis and model interpretation in R.
}

\section{Introduction}\label{introduction}

In machine learning (ML), understanding feature-target relationships is increasingly valued alongside predictive accuracy.
Complex models can capture nonlinear patterns, but are typically considered ``black box'' models due to their opaque internal mechanisms, which has motivated the development of interpretable machine learning (IML) methods \citep{molnar2020interpretable, murdoch2019definitions}.
Among these, feature importance (FI) methods quantify the relevance of input features for a model's predictions, providing insight into which features drive model behavior \citep{murdoch2019definitions, fisher2019allmodelsg}.
The applications range from increasing our understanding of a given data-generating process (DGP) to practical issues such as feature selection \citep{guyon2003introductionvariablea, guidotti2018survey}.

We present \CRANpkg{xplainfi}, an R package that provides a unified interface for computing and comparing global, loss-based FI methods, which measure the change in predictive performance when features are removed, perturbed, or marginalized \citep{ewald2024guidefeatureb}.
Alternative approaches out of scope include local FI methods (i.e., FI values for individual predictions such as Shapley values \citep[see][]{shapley1953value, rozemberczki2022shapleyvalue}) and variance-based sensitivity measures, such as Sobol indices \citep{sobol2001globalsensitivitya}.
When we refer to FI methods from here on, we mean global, loss-based FI methods unless otherwise stated.
Various FI methods are implemented across packages in both the R and Python ecosystems, but in the R ecosystem, some methods are completely absent.
\CRANpkg{xplainfi} aims to fill this gap by providing a larger collection of FI methods than previously available, along with p-values and confidence intervals for FI uncertainty quantification, which are often needed in applications.

The package is built on top of the \CRANpkg{mlr3} ecosystem \citep{lang2019mlr3moderna}, which allows development to focus on the FI methods themselves without re-implementing common building blocks such as abstractions for learning algorithms, resampling, tuning, and pipelines \citep{binder2021mlr3pipelinesflexible}.
Because many real-world datasets exhibit dependent and correlated features, marginal perturbation-based FI can yield misleading attributions by breaking the dependence structure and redistributing importance across correlated predictors \citep{hooker2021unrestrictedpermutationb, nicodemus2010behaviour, debeer2020conditionalpermutation}.
Conditional importance methods address this by assessing performance changes under interventions that preserve feature dependencies \citep{watson2021testingconditional}, yet remain underrepresented in available implementations.
One challenge with these methods is the need for suitable conditional sampling methods, which is why \CRANpkg{xplainfi} provides multiple methods for continuous and mixed data.
Both conditional importance methods and the ability to handle mixed data are motivated by common issues arising in practical statistical work and data analysis, and they are also areas of active research \citep[see also][]{blesch2024conditionalfeature, redelmeier2020explainingpredictive}.
To that end, we offer \CRANpkg{xplainfi} as a comprehensive, extensible tool to support these efforts.

\subsubsection{Contributions}\label{contributions}

\CRANpkg{xplainfi} provides a unified framework for FI methods built on the \CRANpkg{mlr3} ecosystem.
Key contributions include:
1) Implementation of many standard FI methods, including permutation feature importance (PFI) \citep{breiman2001randomforests}, conditional feature importance (CFI) \citep{strobl2008conditionalvariablec, debeer2020conditionalpermutation}, leave-one-covariate-out (LOCO) \citep{lei2018distributionfreepredictivec}, relative feature importance (RFI) \citep{konig2021relativefeatured}, and both marginal and conditional SAGE (Shapley Additive Global importancE) \citep{covert2020understandingglobalb};
2) Seamless integration with \CRANpkg{mlr3}'s learners, tasks, measures, and resampling strategies;
3) A modular conditional sampling interface supporting Gaussian, adversarial random forests (ARF) \citep{watson2023adversarialrandom}, conditional inference trees \citep{hothorn2006unbiasedrecursive}, and knockoff-based samplers \citep{candes2018panninggoldd};
4) The option to target model, learner, or DGP importance, by leveraging \CRANpkg{mlr3}'s support for resampling, ensembling, and tuning;
5) Uncertainty quantification via variance-corrected confidence intervals \citep{nadeau2003inferencegeneralization}, observation-level LOCO inference \citep{lei2018distributionfreepredictivec}, and the CPI testing framework \citep{watson2021testingconditional}.

The paper is structured as follows: First, we give a brief overview of the implemented global, loss-based FI methods and ways to quantify their uncertainty in Section~\hyperref[sec-fi-methods]{2}.
In Section~\hyperref[sec-related]{3}, we present other R and Python packages that implement FI methods.
Section~\hyperref[sec-intro-xplainfi]{4} gives an introduction to \CRANpkg{xplainfi} and showcases core functionality.
Example scenarios, along with comparisons to existing implementations based on FI values and runtime, are provided in Section~\hyperref[sec-benchmark]{5}.
We conclude with a discussion and outlook of \CRANpkg{xplainfi} in Section~\hyperref[sec-discussion]{6}.

Reproducibility and availability: \CRANpkg{xplainfi} is on CRAN and maintained on \href{https://github.com/mlr-org/xplainfi}{GitHub}.
The \href{https://github.com/slds-lmu/paper_2025_xplainfi_benchmark}{GitHub repository} contains all code required to reproduce the results. The code and results are also included in the online supplement.

\section{Implemented feature importance methods}\label{sec-fi-methods}

We briefly review the FI methods implemented in \CRANpkg{xplainfi}, drawing on the comprehensive overview in \citet{ewald2024guidefeatureb}, to which we refer for a broader discussion and additional references.
We also summarize the statistical inference procedures supported by the package.
Rather than providing a full theoretical treatment, we focus on the estimands and quantities that are required for computation and implementation.

\textbf{Notation}: We assume the supervised learning setting with \(n\) observations of \(p\) features \(X = (X_1, \ldots, X_p)\) and a target \(Y\).
A model \(\hat{f}\) is trained to predict \(Y\) from \(X\), and its performance is evaluated using a loss function \(L(Y, \hat{f}(X))\).
The \emph{feature of interest} (FOI) is denoted \(X_j\), with \(X_{-j}\) representing all remaining features.
More generally, \(X_S\) denotes a subset of features indexed by \(S \subseteq \{1, \ldots, p\}\).
Feature importance for feature \(j\) is denoted \(\text{FI}_j\), with specific methods indicated by subscripts (e.g., \(\text{PFI}_j\), \(\text{LOCO}_j\)).

One central distinction between FI methods is whether they target \emph{marginal} (unconditional) association between a feature and the target, \(X_j \not\!\perp\!\!\!\perp Y\), or \emph{conditional} association given a set of features \(G\), \(X_j \not\!\perp\!\!\!\perp Y \mid X_G\) \citep{strobl2008conditionalvariablec, watson2021testingconditional}.
Many well-established model-independent methods target marginal association (e.g., correlation- or information-theoretic measures) \citep{li2018featureselection, bommert2020benchmarkfilter}.

In the paper and \CRANpkg{xplainfi}, we primarily focus on conditional association of \(X_j\) with \(Y\), often with the important special case of conditioning on all remaining features \(X_{-j}\).
This captures the \emph{incremental} predictive value of a feature when the others are already available, and it entails a substantially harder estimation problem than marginal association, typically requiring predictive models and procedures that respect the joint structure of the features.
Conditioning on arbitrary feature subsets is also supported by \CRANpkg{xplainfi}, e.g., via relative feature importance (RFI) (see Section~\hyperref[sec-fi-perturbation]{2.2}).
Importantly, the loss-based FI methods considered below quantify \emph{predictive} importance (or reliance) under a particular intervention scheme (e.g., permutation, conditional resampling, refitting, or marginalization) and are therefore not equivalent to generic measures of statistical dependence.
For certain population-level (oracle) importance parameters---including refitting-based measures in the framework of \citet{williamson2023generalframeworkb} under suitable losses---a zero importance parameter can be equivalent to conditional independence, but FI does not imply a causal effect of \(X_j\) on \(Y\), and estimated FI values can depend on the chosen intervention distribution as well as the presence of correlated or redundant predictors.

\subsection{Estimands and statistical inference}\label{sec-fi-estimands}

There are three possible targets of an FI analysis, corresponding to distinct estimands:
a fitted \emph{model}, a given \emph{learner}, or the true prediction function regarding the DGP, also called population level inference \citep[see also][]{chiaburu2024uncertainty, molnar2023relatingpartialc}.
\emph{Model importance} analyzes a single, fixed model and quantifies how it uses features for prediction.
\emph{Learner importance} targets the learning algorithm, and the model it produces, when training data is sampled from the DGP of a given sample size.
It usually requires refitting models via resampling to measure importance across multiple model instantiations, thereby capturing how much models of this class rely on each feature.
\emph{DGP importance} aims to explain the true feature-target relationship, by analyzing the population level prediction function (or Bayes-optimal predictor), e.g., \(f_0(x) = \mathbb{E}[Y \mid X = x]\) under \(\ell_2\) loss.
As we usually never have direct access to this Bayes-optimal predictor in practical applications, this function must then be approximated by a strong learner in an AutoML-like fashion, e.g., by optimizing over multiple model classes and hyperparameters \citep{thornton2013autoweka} or by constructing heterogenous stacking ensembles \citep{erickson2020autogluontabular, vanderlaan2007superlearner}.
In practice this is very similar to \emph{Learner importance}, while model building is usually more expensive.

Statistical inference can be performed at each of the three estimand levels.
At the model level, hypothesis tests can be based on observation-wise losses on a single test set.
At the learner level, variance-corrected tests can be applied to paired loss differences across resampling iterations.
Since FI estimation inherently involves random components (e.g., from data resampling, permutation, or stochastic learners), repeated evaluation is generally needed to quantify this uncertainty, just as cross-validation is needed for reliable performance estimation.
\CRANpkg{xplainfi} provides dedicated inference methods at the model and learner levels, which therefore also includes
population level inference.

When evaluating multiple FOIs simultaneously, any of these inference procedures introduces a multiple-testing problem.
Depending on the use case, it is necessary to control either the family-wise error rate (FWER) via methods such as the Bonferroni-Holm correction, or to control the false-discovery rate (FDR) with methods such as the Benjamini-Hochberg or Benjamini--Yekutieli procedures, where the latter is valid under arbitrary dependence structures \citep{holm1979simplesequentially, benjamini1995controllingfalsea, benjamini2001controlfalse}.
In an exploratory setting, it is often sufficient to control the FDR, which controls the expected proportion of false positives among all rejections (e.g., 1 out of 20 features deemed important), whereas controlling the FWER is more suitable for confirmatory settings in which avoiding false positives takes priority.

Regardless of the estimand, loss-based FI methods are usually constructed by ``removing'' the information about the FOI(s) and calculating the difference in expected loss (with vs.~without this information).
Following \citet{ewald2024guidefeatureb}, we group the FI methods by the strategy used to remove information: feature perturbations, model refitting, or Shapley values.

\subsection{Methods based on feature perturbations}\label{sec-fi-perturbation}

The most well-known method in this category is \emph{permutation feature importance} (PFI), originally introduced by \citet{breiman2001randomforests} for random forests.
PFI targets the effect of breaking the link between \(X_j\) and the rest of the joint distribution at evaluation time, holding \(\hat{f}\) fixed.
The model is fit once but evaluated twice: once with the original features and once with the FOI \(X_j\) replaced by a perturbed version \(\tilde{X}_j\).
In PFI, \(\tilde{X}_j\) is derived by simply shuffling \(X_j\) in-place, which keeps the marginal distribution of \(X_j\) intact.
Importance is then measured as the difference in performance between the original model prediction and the model prediction using the uninformative replacement for the FOI.
Formally,

\[
\text{PFI}_j = \mathbb{E}\left[L\left(Y, \hat{f}(\tilde{X}_j, X_{-j})\right)\right] - \mathbb{E}\left[L\left(Y, \hat{f}(X)\right)\right].
\]

While PFI is computationally cheap, requiring only model predictions and simple shuffling of feature vectors, it breaks dependencies not only between the FOI and the target, but also between the FOI and all other features.
This can yield implausible feature combinations and misleading attributions under feature dependence \citep[see][]{hooker2021unrestrictedpermutationb}.
\emph{Conditional permutation feature importance} (CFI), introduced by \citet{strobl2008conditionalvariablec}, addresses this by perturbing the FOI conditional on a set of other features: \(X_j\) is replaced by \(\tilde{X}_j\) sampled from the conditional distribution of \(X_j \mid X_{-j}\), thereby preserving (parts of) the dependence structure among the predictors.
\citet{konig2021relativefeatured} generalize PFI and CFI to \emph{relative feature importance} (RFI) by specifying a conditioning set \(G\) for which \(\tilde{X}_j\) retains conditional dependencies, i.e., \(\tilde{X}_j\) is sampled from the conditional distribution of \(X_j \mid X_G\); this yields CFI for \(G = -j\) and PFI for \(G = \emptyset\).

Since the permutation (or sampling) step introduces randomness, multiple repetitions should be used to stabilize the estimate of the importance score.
In \CRANpkg{xplainfi}, this is controlled via the \texttt{n\_repeats} parameter, with the final importance score averaged across repetitions.
In terms of inference, \CRANpkg{xplainfi} allows applying the corrected t-test approach proposed by \citet{nadeau2003inferencegeneralization} to paired loss differences across resampling iterations, which can be used to quantify uncertainty of (learner-level) FI estimates based on repeated data splits.
\citet{molnar2023relatingpartialc} evaluate and recommend this approach for PFI in combination with subsampling or bootstrapping with 10--15 iterations.
\citet{schulz-kumpel2025constructingconfidence} recommend for this test a train-test ratio of 0.9 and 25 subsampling iterations.
\CRANpkg{xplainfi} lets the user apply the inference method to all FI methods, while warning the user if fewer than 10 bootstrap or subsampling iterations are used.

Beyond conditional permutation-based approaches, the conditional predictive impact (CPI) framework proposed by \citet{watson2021testingconditional} provides a dedicated hypothesis test for conditional feature importance.
\CRANpkg{xplainfi} offers both the original version based on the knockoff framework \citep{candes2018panninggoldd} and an ARF-based version for mixed data \citep{blesch2025conditionalfeaturea}.

The conditional sampling required by CFI, RFI, and conditional SAGE (described below) can be performed using different approaches, each with distinct trade-offs.
The Gaussian approach assumes multivariate normality, making it fast and straightforward but limited to continuous features.
Conditional inference trees provide a nonparametric alternative that handles mixed feature types, as proposed by \citet{redelmeier2020explainingpredictive} for conditional distribution estimation (e.g., for conditional Shapley-style estimands) and implemented in \CRANpkg{shapr}.
Adversarial random forests offer flexible density estimation and conditional sampling for mixed data at a higher computational cost \citep{watson2023adversarialrandom}.
Finally, knockoff sampling provides a specialized approach that enables valid inference through the CPI framework, but implementations for mixed data are less common \citep{candes2018panninggoldd, blesch2024conditionalfeature}.
Gaussian, conditional inference tree, and ARF samplers can be applied to all conditional importance methods in \CRANpkg{xplainfi}, while knockoff sampling is only compatible with CFI.

\subsection{Methods based on model refitting}\label{sec-fi-refit}

A more intuitive approach to remove an FOI's information is to refit a model without the FOI (using the same learner) and then measure the performance difference between the reduced and the original (full) model.
The best-known method following this approach is the \emph{leave-one-covariate-out} (LOCO) \citep{lei2018distributionfreepredictivec}, which analyzes a single FOI \(X_j\).
If one is interested in the importance of a set of features at once,
this is known as \emph{leave-one-group-out} (LOGO) \citep{au2022grouped}.
\citet{williamson2023generalframeworkb} proposed a general framework for refitting-based methods, including a dedicated statistical inference method.
In \CRANpkg{xplainfi}, we use the terms LOCO and WVIM (\emph{Williamson's variable importance measure}; \citet{ewald2024guidefeatureb} proposed the acronym).

This approach is conceptually simple: To evaluate an FOI's importance, the model in question is fit twice, once with and once without the FOI.
Again, the resulting performance difference gives a straightforward indication of the predictive value ``lost'' by leaving out feature \(X_j\):

\[
\text{LOCO}_j = \mathbb{E}\left[L\left(Y, \hat{f}_{-j}(X_{-j})\right)\right] - \mathbb{E}\left[L\left(Y, \hat{f}(X)\right)\right],
\]

where \(\hat{f}_{-j}\) denotes the model fitted without feature \(X_j\).
Model refitting, of course, increases the computational cost in comparison to FI methods based on perturbations, where only one model needs to be fit as the basis for all subsequent calculations.
This can be problematic when using LOCO (or a generalization thereof) on datasets with many features, especially when combined with expensive learners.
When using stochastic learners, it is also recommended to perform multiple refits (in \CRANpkg{xplainfi} via \texttt{n\_repeats}) to obtain stable importance estimates, further increasing the required computation time.

The inverse operation, \emph{leave-one-covariate-in} (LOCI), trains models with only a single feature and compares performance against a featureless baseline to measure the increase in performance.
While theoretically consistent, LOCI is rarely useful in practice as it essentially investigates univariate associations between the target and a single feature, for which more appropriate methods are available (see above).
The group version of LOCI (also included in the WVIM framework) is more useful, where a subset of the features is ``left in'', chosen, for example, based on domain knowledge about the data.

For statistical inference on LOCO, \citet{lei2018distributionfreepredictivec} suggest an observation-level nonparametric test using the \(\ell_1\) loss differences on a test set, which is implemented in \CRANpkg{xplainfi} in a generalized manner that also enables other losses and tests.
The corrected t-test approach mentioned in the previous section is also available for LOCO, but it has not been explicitly investigated in this context.

\subsection{Methods based on Shapley values}\label{sec-fi-shapley}

Shapley values originate in cooperative game theory \citep[see][]{shapley1953value} and, in ML, are often used for ``local'' explanations, i.e., for explaining individual model predictions for selected data instances.
But several methods have also been proposed to apply the underlying principle to global FI, including SFIMP (\emph{Shapley Feature IMPortance}) by \citet{casalicchio2019visualizingfeatureb} and SPVIM (\emph{Shapley Population Variable Importance Measure}) by \citet{williamson2020efficientnonparametrica}, which is related to the more widely adopted SAGE (\emph{Shapley Additive Global importancE}) by \citet{covert2020understandingglobalb}.
\CRANpkg{xplainfi} implements SAGE, which distributes the overall model loss across the features based on their individual contributions, yielding importance scores that sum to the overall model loss.
The SAGE value for feature \(X_j\) is defined as the weighted average of its marginal contributions to feature subsets \(S\), also called coalitions, based on a value function \(v\):
\[
\text{SAGE}_j = \sum_{S \subseteq P \setminus \{j\}} \frac{|S|!(p-|S|-1)!}{p!} \left[ v(S \cup \{j\}) - v(S) \right].
\]

Here, \(v(S)\) measures how much better the model performs when the information of the features in \(S\) is used for prediction (while features outside \(S\) are integrated out) compared to a featureless baseline.
\emph{Marginal SAGE} (mSAGE) uses the marginal distribution, resulting in

\[
v_m(S) = \mathbb{E}\left[L(Y, \hat{f}_\emptyset)\right] - \mathbb{E}\left[L\left(Y, \mathbb{E}_{X_{-S}}\left[\hat{f}(X_S, X_{-S})\right]\right)\right],
\]

and \emph{conditional SAGE} (cSAGE) uses the conditional distribution, replacing \(\mathbb{E}_{X_{-S}}\) with \(\mathbb{E}_{X_{-S}|X_S}\) in the equation above to yield \(v_c(S)\).

Since exact computation requires evaluating all possible feature coalitions (\(2^p\)), implementations must rely on approximations.
\CRANpkg{xplainfi} implements the permutation estimator described by \citet{covert2020understandingglobalb}, where the feature coalitions are built from one of \texttt{n\_permutations} shuffles of the feature vector.
The empty coalition is always evaluated, resulting in a total number of evaluated coalitions of \(1 + p \cdot \texttt{n\_permutations}\).
The variance of SAGE value estimates along the evaluated permutations is also used to define an ``early stopping'' criterion to avoid excessive computations.
\citet{covert2020understandingglobalb} use these variances to construct confidence-like intervals, but we do not include them as statistical inference methods because their coverage is unclear.
Additionally, the marginalization step of the estimation requires a subset of size \texttt{n\_samples} for each data instance, and larger values yield more stable SAGE value estimates but incur additional time and memory costs for large datasets.

\section{Related work}\label{sec-related}

Various R packages on CRAN implement one or more FI methods.
We compare them along two dimensions:
(1) the \emph{scope of FI methods} they provide, and
(2) whether they support \emph{model importance} (analyzing a single pre-fit model) or also \emph{learner importance} via resampling and refitting (see Section~\hyperref[sec-fi-methods]{2}).

\CRANpkg{vip} \citep{greenwell2020variable} requires a pre-fit model and provides model-specific importance extraction, PFI, a variance-based method (``FIRM''), and a Shapley-based measure based on the mean absolute Shapley/SHAP values per feature, i.e., an aggregate of per-observation feature attributions of the model prediction.
Unlike loss-based FI methods such as PFI and SAGE, both FIRM and the Shapley-based importance in \CRANpkg{vip} summarize importance on the model-output (prediction) scale rather than via changes in loss.
It supports a wide range of model classes and offers repeated permutations and optional subsampling of the evaluation data, but does not refit models;
accordingly, it targets model importance.
Similarly, \CRANpkg{iml} (no longer actively developed) operates on pre-fit models via a \texttt{Predictor} wrapper and provides PFI alongside other IML methods such as feature effects, interaction statistics, and local explanations.
\CRANpkg{DALEX} likewise wraps pre-fit models via an explainer object;
its PFI implementation is provided by the companion package \CRANpkg{ingredients}, complemented by local explanations and feature-effect visualizations.
\CRANpkg{hstats} provides PFI alongside H-statistics for interaction detection, also operating on pre-fit models.
\CRANpkg{flashlight} wraps pre-fit models in an explainer abstraction and provides model-agnostic PFI, interaction-strength measures based on Friedman's \(H\)-statistics for interaction detection, as well as Shapley-based importance via mean absolute (approximate) SHAP feature attributions aggregated across observations.
It supports side-by-side comparison of multiple fitted models via the \texttt{multiflashlight} function, which can also be used to compare models that were manually refit on different resampling splits.
However, it does not orchestrate refitting/resampling or aggregate fold-wise importance into a single learner importance estimate.
The \CRANpkg{mlr3} ecosystem includes \CRANpkg{mlr3filters}, which provides PFI with resampling support within its feature-selection framework, but is designed for feature ranking rather than importance analysis with uncertainty quantification.
Generally, PFI is also available in many model-specific implementations (e.g., in \CRANpkg{ranger} and \CRANpkg{randomForestSRC} for random forests).

To the best of our knowledge, there is no general-purpose, model-agnostic R implementation of CFI or RFI that mirrors the kind of model-wrapping interfaces available for PFI.
\CRANpkg{cpi} implements Conditional Predictive Impact (CPI), a dedicated conditional-importance testing framework based on knockoff sampling \citep{watson2021testingconditional}.
It is tied to the knockoff construction rather than providing a general conditional-sampling interface for CFI/RFI and focuses on hypothesis testing (p-values and confidence intervals) rather than a broad FI computation framework.
\CRANpkg{permimp} implements a version of CFI but is restricted to tree-based methods and is therefore not model-agnostic.

\CRANpkg{vimp} takes a different approach: rather than evaluating a pre-fit model, it refits learners for each evaluated feature subset using the \CRANpkg{SuperLearner} framework, which automatically ensembles prediction models to approximate the Bayes-optimal predictor \citep{williamson2023generalframeworkb}.
This targets population-level importance with valid confidence intervals and hypothesis tests.
\footnote{But note that the \CRANpkg{SuperLearner} interface also allows using only a single learner, which can then be analyzed for learner importance with \CRANpkg{vimp}}
It implements what it calls ``conditional VIM'' (equivalent to LOCO for individual FOIs), ``marginal VIM'' (equivalent to LOCI), and SPVIM (\emph{Shapley Population Variable Importance Measure}; \citet{williamson2020efficientnonparametrica}).
SPVIM is a Shapley-based method that, like SAGE, distributes the overall model performance across features according to their marginal contributions.
Unlike SAGE, it replaces the marginalization or conditional sampling step with model refitting for each feature coalition, which can be slow, but uses Kernel SHAP \citep{covert2021improvingkernelshap} to approximate the Shapley values.
\CRANpkg{vimp} also accepts pre-computed prediction vectors, which, in principle, allow a form of model importance, but with a notably different API compared to the model-wrapping approach of the other packages.
It is tightly coupled with the \CRANpkg{SuperLearner} framework and offers less flexibility in resampling strategies.

In the Python ecosystem, \texttt{scikit-learn} provides tree-based PFI and model-agnostic PFI, both operating on pre-fit estimators.
The \texttt{ELI5} project similarly provides PFI as its only model-agnostic FI method \citep{eli5debug}.
\texttt{sage} offers the original (marginal) SAGE implementation with no dedicated interface for conditional sampling, but it allows supplying external surrogate models for the same purpose \citep{covert2020understandingglobalb, covert2021improvingkernelshap}.
The \texttt{fippy} package in Python offers the closest overlap with \CRANpkg{xplainfi} in terms of perturbation-based methods, providing PFI, CFI, RFI, and both marginal and conditional SAGE using various sampling methods.
It also supports simple Wald-type CIs for FI scores from the observation-based losses on test data if the chosen performance measure is decomposable.
Overall, \texttt{fippy} is mainly centered around explaining pre-fit estimators on fixed evaluation data, but it also offers a (somewhat underdeveloped) \texttt{LearnerExplainer} which only supports CFI in combination with subsampling.

In summary, \CRANpkg{xplainfi} offers the most comprehensive coverage of model-agnostic FI methods in R, including PFI, CFI, RFI, LOCO, WVIM, and both marginal and conditional SAGE, with flexible systems covering learners, metrics, resampling, and model/learner importance, including tuning and complex ensembles for the latter.
Among the compared R packages, only \CRANpkg{xplainfi} and \CRANpkg{vimp} provide statistical inference methods for their importance estimates,
including confidence intervals and hypothesis tests.
In the Python ecosystem, \texttt{fippy} also provides inference, but limited to the model-importance setting.
Table \ref{tab:tab-related-method} gives a comparative overview.

\begin{table}
\centering
\caption{\label{tab:tab-related-method}Overview of FI methods and capabilities in xplainfi and related R packages and selected Python packages. A checkmark indicates an available feature, and a checkmark in parenthesis indicates a feature is not directly offered as such but available in either a limited or indirect fashion. Model importance refers to analyzing a pre-fit model; learner importance refers to refitting models via resampling. Statistical inference denotes the availability of confidence intervals or hypothesis tests for importance estimates.}
\centering
\resizebox{\ifdim\width>\linewidth\linewidth\else\width\fi}{!}{
\begin{tabular}[t]{lcccccc>{}c|cc}
\toprule
  & xplainfi & vimp & vip & iml & hstats & flashlight & DALEX & fippy & sage\\
\midrule
\addlinespace[0.3em]
\multicolumn{10}{l}{\textbf{Methods}}\\
\hspace{1em}PFI & \ding{51} &  & \ding{51} & \ding{51} & \ding{51} & \ding{51} & \ding{51} & \ding{51} & \\
\hspace{1em}CFI & \ding{51} &  &  &  &  &  &  & \ding{51} & \\
\hspace{1em}RFI & \ding{51} &  &  &  &  &  &  & \ding{51} & \\
\hspace{1em}LOCO & \ding{51} & \ding{51} &  &  &  &  &  &  & \\
\hspace{1em}WVIM & \ding{51} & \ding{51} &  &  &  &  &  &  & \\
\hspace{1em}mSAGE & \ding{51} &  &  &  &  &  &  & \ding{51} & \ding{51}\\
\hspace{1em}cSAGE & \ding{51} &  &  &  &  &  &  & \ding{51} & (\ding{51})\\
\hspace{1em}SPVIM &  & \ding{51} &  &  &  &  &  &  & \\
\addlinespace[0.3em]
\multicolumn{10}{l}{\textbf{Capabilities}}\\
\hspace{1em}Model importance & \ding{51} & \ding{51} & \ding{51} & \ding{51} & \ding{51} & \ding{51} & \ding{51} & \ding{51} & \ding{51}\\
\hspace{1em}Learner importance & \ding{51} & \ding{51} &  &  &  &  &  & (\ding{51}) & \\
\hspace{1em}Statistical inference & \ding{51} & \ding{51} &  &  &  &  &  & \ding{51} & \\
\bottomrule
\end{tabular}}
\end{table}

\section{An introduction to xplainfi}\label{sec-intro-xplainfi}

\CRANpkg{xplainfi}'s architecture is heavily inspired by the \CRANpkg{mlr3} framework and its extension packages, which also form the foundation of the underlying computational infrastructure.
This means primarily two things for the user:
1. The package API is based on \CRANpkg{R6} classes with the corresponding object-oriented design, using \CRANpkg{mlr3} functions and concepts.
2. \CRANpkg{xplainfi}'s basic ML capabilities are determined by the \CRANpkg{mlr3} ecosystem.
We begin with a brief introduction to \CRANpkg{mlr3} and then build from simple to more advanced applications of \CRANpkg{xplainfi}.
For a complete overview, we refer to \citet{mlr3book}, and in particular to \citet{mlr3bookdatabasicmodeling} for a full introduction.

\subsection{\texorpdfstring{A brief introduction to \texttt{mlr3}}{A brief introduction to mlr3}}\label{a-brief-introduction-to-mlr3}

\CRANpkg{mlr3} is an R package and an associated ecosystem of extension packages born from methodological and applied ML research.
Its core components are abstractions for the building blocks of ML pipelines.
Most importantly for our purposes, these include the objects shown in this short application example:

\begin{verbatim}
library(mlr3learners) # loads mlr3
rr <- resample(
  learner = lrn("classif.ranger", num.trees = 100),
  task = tsk("penguins"),
  resampling = rsmp("cv", folds = 5))
rr$score(msr("classif.ce"))
\end{verbatim}

\begin{verbatim}
#>     task_id     learner_id resampling_id iteration classif.ce
#> 1: penguins classif.ranger            cv         1     0.0000
#> 2: penguins classif.ranger            cv         2     0.0290
#> 3: penguins classif.ranger            cv         3     0.0000
#> 4: penguins classif.ranger            cv         4     0.0000
#> 5: penguins classif.ranger            cv         5     0.0294
#> Hidden columns: task, learner, resampling, prediction_test
\end{verbatim}

\begin{verbatim}
rr$aggregate(msr("classif.ce"))
\end{verbatim}

\begin{verbatim}
#> classif.ce 
#>     0.0117
\end{verbatim}

Here we applied the random forest classification learner as implemented in \CRANpkg{ranger} with 100 trees to the \texttt{penguins} dataset \citep{horst2020palmerpenguinspalmer} in a 5-fold cross-validation procedure, evaluated each iteration with the classification error (CE), and finally calculated the average CE across iterations.

A \texttt{Task} (\texttt{tsk()}) encapsulates the data and the learning problem (regression, classification, etc.).
Built-in tasks for common datasets are available, e.g., \texttt{tsk("penguins")}, and custom tasks can be created from any \texttt{data.frame}-like object, e.g., \texttt{as\_task\_regr(mtcars,\ target\ =\ "mpg")}.
A \texttt{Learner} (\texttt{lrn()}) represents an ML algorithm with its hyperparameters.
For example, \texttt{lrn("regr.ranger",\ num.trees\ =\ 100)} creates a random forest learner via \CRANpkg{ranger} using 100 trees.
Learners can also be extended into full ML pipelines with preprocessing, feature extraction, or hyperparameter optimization \citep[see][\citet{mlr3booktuning}, \citet{binder2021mlr3pipelinesflexible}]{mlr3bookpreprocessing}.
A \texttt{Measure} (\texttt{msr()}) quantifies prediction performance, e.g., \texttt{msr("regr.mse")} for MSE for a regression task.
Some measures are decomposable into observation-wise losses, which is relevant for certain inference methods described below.
A \texttt{Resampling} (\texttt{rsmp()}) defines train-test splitting strategies, such as holdout, k-fold CV, bootstrapping, or subsampling, e.g., \texttt{rsmp("cv",\ folds\ =\ 5)}.

The general \CRANpkg{xplainfi} API relies on these components and applies them to all implemented feature importance methods.
For each feature importance method, the user first defines the method object by specifying a \texttt{Task} as the target for the analysis, a \texttt{Learner} for training and predictions, a \texttt{Resampling} strategy, and a \texttt{Measure} for evaluating predictions.
The \texttt{\$compute()} method is then called to perform the actual computational steps required for the individual importance method.
Finally, importance scores can be accessed at different levels of aggregation:
\texttt{\$importance()} returns importance values per feature, aggregated across resampling iterations and repetitions (e.g., permutation repetitions in \texttt{PFI});
\texttt{\$scores()} returns importance values per feature \emph{and} per resampling iteration and repetition, allowing for custom aggregation or visualization;
and \texttt{\$obs\_loss()} returns observation-wise loss scores and importance values, if available for the current importance method and \texttt{Measure}.

In the following, we showcase \CRANpkg{xplainfi} on an included DGP, and refer to \href{https://mlr-org.github.io/xplainfi/}{the package website} for additional tutorials and descriptions of this and other illustrative simulation settings.

\subsection{\texorpdfstring{Example 1: PFI with \CRANpkg{xplainfi}}{Example 1: PFI with }}\label{example-1-pfi-with}

We start by calculating PFI on a synthetic task with four normally distributed features \texttt{x1} through \texttt{x4}, two of which are correlated (\texttt{x1} and \texttt{x2}, \(r = 0.8\)) and two of which are independent (\texttt{x3} and \texttt{x4}).
The DGP is \(y = 2 x_1 + x_3 + \varepsilon\), (\(\varepsilon \sim \mathcal{N}(0, 0.04)\)), generated by the \texttt{sim\_dgp\_correlated()} simulation utility function included in the package, which produces a \texttt{Task} object.

\subsubsection{Model importance}\label{model-importance}

The simplest use case is analyzing a single pre-trained model.
We train a \CRANpkg{ranger} random forest on a holdout split and compute PFI on the corresponding test set:

\begin{verbatim}
library(xplainfi)
task <- sim_dgp_correlated(n = 5000, r = 0.8)
lrn_ranger <- lrn("regr.ranger")
resampling_ho <- rsmp("holdout")$instantiate(task)
lrn_ranger$train(task, row_ids = resampling_ho$train_set(1))

pfi_model <- PFI$new(
  task = task, learner = lrn_ranger, measure = msr("regr.mse"),
  resampling = resampling_ho, n_repeats = 10)
pfi_model$compute()
pfi_model$importance()
\end{verbatim}

\begin{verbatim}
#>    feature importance
#> 1:      x1   6.67e+00
#> 2:      x2   1.52e-01
#> 3:      x3   1.82e+00
#> 4:      x4   2.02e-05
\end{verbatim}

When a pre-trained learner is passed, \CRANpkg{xplainfi} detects this automatically and skips the training step, using the fitted model directly for prediction.
The resampling must be instantiated with exactly one test set (i.e., holdout), as there is only one model to evaluate.
This yields \emph{model importance}: the importance scores reflect the behavior of this specific model on this specific test set.

\subsubsection{Learner importance}\label{learner-importance}

To capture importance at the \emph{learner} level, we pass an untrained learner together with a resampling strategy.
\CRANpkg{xplainfi} then trains a new model in each resampling iteration, and importance scores are aggregated across iterations:

\begin{verbatim}
pfi <- PFI$new(
  task = task, learner = lrn("regr.ranger"), measure = msr("regr.mse"),
  resampling = rsmp("cv", folds = 3), n_repeats = 10)
pfi$compute()
pfi$importance()
\end{verbatim}

\begin{verbatim}
#>    feature importance
#> 1:      x1   6.535166
#> 2:      x2   0.163446
#> 3:      x3   1.813342
#> 4:      x4  -0.000126
\end{verbatim}

The construction of the \texttt{PFI} object and the computation are separated: calling \texttt{\$new()} defines the setup, while \texttt{\$compute()} performs the actual work.
To access individual importance scores for each permutation and resampling iteration, the \texttt{\$scores()} method is available, which stores the corresponding feature, iteration indices, and the associated measure value of the original (baseline) model and the model post perturbation:

\begin{verbatim}
head(pfi$scores(), 3)
\end{verbatim}

\begin{verbatim}
#>    feature iter_rsmp iter_repeat regr.mse_baseline regr.mse_post importance
#> 1:      x1         1           1            0.0695          6.50       6.43
#> 2:      x1         1           2            0.0695          6.46       6.39
#> 3:      x1         1           3            0.0695          6.68       6.61
\end{verbatim}

Using resampling not only captures learner-level variation but also enables variance-corrected inference methods such as the Nadeau-Bengio correction (see Example 3), which require loss differences across resampling iterations and are therefore not available for model importance.

All other components are easily swappable.
Here we rerun PFI with XGBoost, \(R^2\), and subsampling instead:

\begin{verbatim}
pfi <- PFI$new(
  task = task,
  learner = lrn("regr.xgboost", eta = 0.01, nrounds = 2000),
  measure = msr("regr.rsq"),
  resampling = rsmp("subsampling", repeats = 10),
  n_repeats = 15)
pfi$compute()
pfi$importance()
\end{verbatim}

\begin{verbatim}
#>    feature importance
#> 1:      x1   1.62e+00
#> 2:      x2   1.08e-03
#> 3:      x3   3.81e-01
#> 4:      x4   6.75e-05
\end{verbatim}

\CRANpkg{xplainfi}'s capabilities are extendable by the \CRANpkg{mlr3} ecosystem: any learner from \CRANpkg{mlr3learners} or \texttt{mlr3extralearners} \citep{fischer2025mlr3extralearnersexpanding} can be used, including ``auto-tuned'' learners, pipelines via \CRANpkg{mlr3pipelines} or full AutoML systems \citep{mlr3bookadvancedtuning, mlr3bookpipelinestuning}.
Furthermore, any metric from \CRANpkg{mlr3measures} is available.
The trained models and baseline results are retained via \texttt{\$resample\_result} for further analysis.

\subsection{Example 2: CFI and conditional samplers}\label{example-2-cfi-and-conditional-samplers}

As discussed in Section~\hyperref[sec-fi-methods]{2}, PFI can produce misleading results when features are correlated.
CFI addresses this by sampling from the conditional distribution, which requires a \emph{conditional sampler}.
\CRANpkg{xplainfi} provides a modular abstraction for this.
Several conditional samplers are available, but here we focus on two:
The \texttt{ConditionalGaussianSampler} assumes multivariate normality and is fast but limited to numeric features.
The \texttt{ConditionalARFSampler} uses adversarial random forests (ARF) and handles mixed data, but is computationally more expensive \citep{watson2023adversarialrandom}.

Samplers are instantiated on a given task once, after which one can draw one or more observations, which \texttt{CFI} and related methods handle internally.
Each conditional sampler allows the specification of an arbitrary conditioning set of features for sampling.

We continue with the correlated DGP from Example 1, where \texttt{x1} and \texttt{x2} are correlated (\(r = 0.8\)), but only \texttt{x1} affects the target.
Using a Gaussian conditional sampler, we compute CFI and also request quantiles calculated across resampling iterations:

\begin{verbatim}
cfi = CFI$new(
  task = task, learner = lrn("regr.ranger"), measure = msr("regr.mse"),
  resampling = rsmp("subsampling", repeats = 5),
  sampler = ConditionalGaussianSampler$new(task),
  n_repeats = 10)
cfi$compute()
cfi$importance(ci_method = "quantile", alternative = "two.sided")
\end{verbatim}

\begin{verbatim}
#>    feature importance conf_lower conf_upper
#> 1:      x1   2.78e+00   2.724271   2.828378
#> 2:      x2  -7.67e-04  -0.002407   0.000507
#> 3:      x3   1.81e+00   1.764661   1.878462
#> 4:      x4  -4.11e-05  -0.000561   0.000534
\end{verbatim}

Compared to PFI, CFI identifies that \texttt{x2} is not associated with \texttt{y} conditional on the other features.
The \texttt{sampler} argument distinguishes CFI from PFI: while PFI uses marginal permutation, CFI requires a conditional sampler.
The \texttt{"quantile"} method shown above provides empirical quantiles from the distribution of importance scores across resampling iterations, which help to gauge the stability of the estimates.
Note that via \texttt{"alternative"}, we explicitly requested the default of \texttt{"two-sided"} intervals, by analogy with two- or one-sided hypothesis tests, where \texttt{"alternative\ =\ \textquotesingle{}greater\textquotesingle{}"} would have given us only the 95\% quantile as lower bound.

\subsection{Example 3: Inference}\label{example-3-inference}

For principled inference beyond empirical quantiles, \CRANpkg{xplainfi} supports two main approaches, depending on the chosen importance method.

\subsubsection{Nadeau-Bengio correction for PFI}\label{nadeau-bengio-correction-for-pfi}

\citet{molnar2023relatingpartialc} recommend variance correction based on \citet{nadeau2003inferencegeneralization} when using PFI with subsampling or bootstrapping.
The correction accounts for dependence between resampling iterations that share training observations, yielding confidence intervals with improved (though still imperfect) coverage.
The recommended setup uses approximately 15 subsampling iterations.
Since our earlier PFI example already uses subsampling, we can request corrected confidence intervals via \texttt{ci\_method\ =\ "nadeau\_bengio"}:

\begin{verbatim}
head(pfi$importance(ci_method = "nadeau_bengio"), 3)
\end{verbatim}

\begin{verbatim}
#>    feature importance       se statistic  p.value conf_lower conf_upper
#> 1:      x1    1.62071 0.028601     56.67 8.36e-13   1.556005    1.68541
#> 2:      x2    0.00108 0.000278      3.89 3.67e-03   0.000453    0.00171
#> 3:      x3    0.38069 0.009970     38.18 2.88e-11   0.358136    0.40324
\end{verbatim}

\begin{figure}

{\centering \includegraphics[width=0.95\linewidth,alt={Dot and whisker plot showing point estimates and error bars of the different methods, with the unadjusted being the narrowest and quantile intervals slightly narrower than the Nadeau & Bengio adjusted intervals being much wider.}]{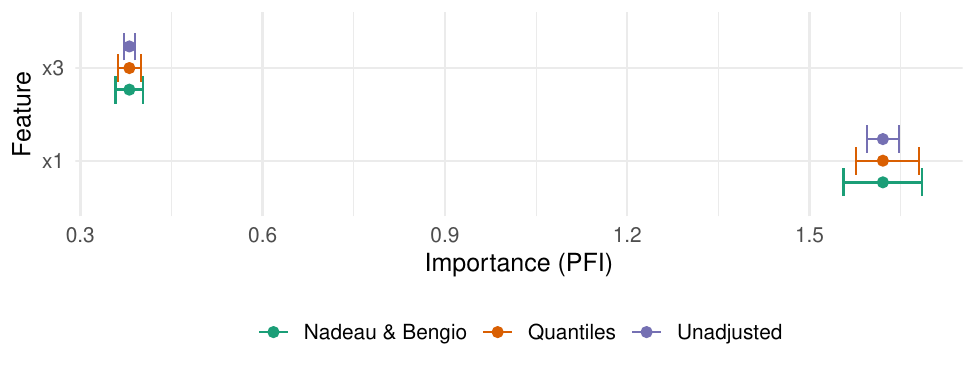} 

}

\caption{Comparison of uncertainty quantification methods using PFI on the correlated task (omitting noise features). Empirical 95\% quantiles fall between the very narrow unadjusted confidence intervals and the wider Nadeau-Bengio-corrected intervals, which show the uncertainty masked by the unadjusted method.}\label{fig:pfi-inference-comparison-plot}
\end{figure}

The output includes standard errors and adjusted confidence intervals.
Note that this approach assumes normally distributed importance scores and was primarily evaluated for PFI; its use with other methods is experimental.
Figure \ref{fig:pfi-inference-comparison-plot} compares the corrected confidence intervals with the empirical 95\% quantiles and also unadjusted confidence intervals.
We note that the latter are also available via \texttt{ci\_method\ =\ "raw"} for comparison, but they are not valid for inference.

\subsubsection{CPI for conditional importance}\label{cpi-for-conditional-importance}

For CFI, inference is available through the CPI framework when using knockoff-based sampling, which leverages observation-wise losses to perform hypothesis tests for feature importance \citep{watson2021testingconditional}.
Using a knockoff sampler with CFI enables \texttt{ci\_method\ =\ "cpi"}, for example in conjunction with a t-test:

\begin{verbatim}
cfi_knockoff = CFI$new(
  task = task, learner = lrn("regr.ranger"), measure = msr("regr.mse"),
  sampler = KnockoffGaussianSampler$new(task),
  resampling = rsmp("holdout"), n_repeats = 1)
cfi_knockoff$compute()
cfi_knockoff$importance(
  ci_method = "cpi", alternative = "greater", 
  p_adjust = "BH", test = "t")
\end{verbatim}

\begin{verbatim}
#>    feature importance       se statistic   p.value conf_lower conf_upper
#> 1:      x1   3.061471 0.101382    30.197 3.88e-160   2.894620        Inf
#> 2:      x2   0.003612 0.002603     1.388  1.10e-01  -0.000672        Inf
#> 3:      x3   1.865136 0.061402    30.376 2.37e-161   1.764083        Inf
#> 4:      x4   0.000148 0.000569     0.261  3.97e-01  -0.000787        Inf
\end{verbatim}

This CPI test provides p-values and one-sided confidence bounds for each feature.
Here we use the \texttt{p\_adjust} argument to use R's \texttt{p.adjust()} function to correct for the FDR with the Benjamini-Hochberg procedure.
We note that this only corrects the p-values, but not the confidence bounds, which would only be adjusted for \texttt{p\_adjust\ =\ "bonferroni"}.
Alternatively to the knockoff approach, \texttt{ConditionalARFSampler} can be used in place of knockoffs, enabling CPI-style inference for mixed data types as proposed by \citet{blesch2025conditionalfeaturea}, though at a higher computational cost.
Note that CPI was proposed in combination with a fixed model and a test set, whereas \CRANpkg{xplainfi} also allows its use with cross-validation.

\subsection{Example 4: LOCO and WVIM}\label{example-4-loco-and-wvim}

LOCO is implemented as a special case of the more general WVIM framework as described in Section~\hyperref[sec-fi-methods]{2}.
We apply it on the simulated task from Example 1 using a single refit iteration and show the same Bonferroni-adjusted p-values and confidence intervals based on the Wilcoxon test proposed by \citet{lei2018distributionfreepredictivec}:

\begin{verbatim}
loco = LOCO$new(
  task = task, learner = lrn("regr.ranger"), n_repeats = 1,
  measure = msr("regr.mse"), resampling = rsmp("holdout"))
loco$compute()
loco$importance(
  ci_method = "lei", test = "wilcox", 
  p_adjust = "bonferroni")[, c(1:2, 5:7)]
\end{verbatim}

\begin{verbatim}
#>    feature importance   p.value conf_lower conf_upper
#> 1:      x1    0.75309 7.59e-243    1.05107    1.28582
#> 2:      x2    0.00229  1.92e-08    0.00274    0.00732
#> 3:      x3    0.49751 4.44e-224    0.66464    0.82465
#> 4:      x4    0.00409  1.97e-33    0.00614    0.00997
\end{verbatim}

\texttt{WVIM} supports arbitrary feature groups via the \texttt{groups} argument, which accepts a named list specifying features belonging to each group (i.e., LOGO):

\begin{verbatim}
groups = list(correlated = c("x1", "x2"), independent = c("x3", "x4"))
wvim = WVIM$new(
  task = task, learner = lrn("regr.ranger"),
  groups = groups, direction = "leave-out",
  measure = msr("regr.mse"), resampling = rsmp("holdout"),
  n_repeats = 2)
wvim$compute()
wvim$importance()
\end{verbatim}

\begin{verbatim}
#>        feature importance
#> 1:  correlated       4.49
#> 2: independent       1.05
\end{verbatim}

Here, the \texttt{"correlated"} group contains features \texttt{x1} and \texttt{x2}, which are removed together.
WVIM also supports a \texttt{"leave-in"} direction, which trains models with \emph{only} the specified features and compares against a featureless baseline (i.e., LOCI or ``LOGI'').
The \texttt{groups} argument is also available for PFI, CFI, and RFI, where the specified groups of features are then always perturbed at once.

\subsection{Conditional samplers}\label{conditional-samplers}

As described in Section~\hyperref[sec-fi-methods]{2}, conditional methods like CFI and cSAGE require a mechanism to sample from the conditional distribution \(P(X_j | X_{-j})\).
\CRANpkg{xplainfi} provides a modular sampler architecture that allows different sampling strategies to be instantiated and passed to the importance method.
All samplers inherit from the \texttt{FeatureSampler} base class and share a common interface: Instantiate a sampler on a \texttt{Task}, then use the \texttt{\$sample()} method to draw new values for specified features conditional on the remaining ones or a subset of features specified as \texttt{conditioning\_set}.

We demonstrate the ARF sampler on the \texttt{penguins} task, which contains both numeric and categorical features.
After instantiation, we sample new values for the \texttt{body\_mass} feature for a few observations, conditional on all other features (including the categorical \texttt{island}):

\begin{verbatim}
task_penguins = tsk("penguins")
sampler_arf = ConditionalARFSampler$new(task_penguins)
# Original values for comparison
task_penguins$data(
  rows = c(1, 20, 40),
  cols = c("island", "bill_length", "body_mass"))
\end{verbatim}

\begin{verbatim}
#>       island bill_length body_mass
#> 1: Torgersen        39.1      3750
#> 2: Torgersen        46.0      4200
#> 3:     Dream        39.8      4650
\end{verbatim}

\begin{verbatim}
# Sample new body_mass values conditional on all other features
sampler_arf$sample(feature = "body_mass", row_ids = c(1, 20, 40))[, c(6, 3, 4)]
\end{verbatim}

\begin{verbatim}
#>       island bill_length body_mass
#> 1: Torgersen        39.1      4298
#> 2: Torgersen        46.0      3850
#> 3:     Dream        39.8      3989
\end{verbatim}

The sampler produces new values for \texttt{body\_mass} drawn from the estimated conditional distribution \(P(\texttt{body\_mass} | X_{-\texttt{body\_mass}})\).
These sampled values replace the original feature when computing CFI or cSAGE, allowing importance to be assessed while preserving dependencies with other features.

The choice of sampler involves trade-offs between flexibility and computational cost.
The ARF sampler handles mixed data types without making distributional assumptions but requires fitting an adversarial random forest, which increases computational time and memory usage.

\section{Benchmark experiments}\label{sec-benchmark}

To evaluate \CRANpkg{xplainfi}, we performed benchmark experiments in two categories:
\textbf{Importance results}: Since \CRANpkg{xplainfi} re-implements methods available in other packages, we verify that it produces equivalent importance scores across shared methods and DGPs, establishing that implementations are faithful and that results are comparable across packages.
\textbf{Runtime}: We assess whether \CRANpkg{xplainfi}'s implementations are competitive in terms of computational cost compared to existing single-method packages, across varying task dimensionalities and method parameters.

We compare against packages that implement the same methods in a model-importance setting, as this is the common denominator across implementations (see Section~\hyperref[sec-related]{3}).
For PFI, we compare with \CRANpkg{iml} and \CRANpkg{vip}.
For marginal SAGE, we compare with \texttt{fippy} and \texttt{sage}, the latter implementing kernel SAGE \citep{covert2021improvingkernelshap}.
For CFI and conditional SAGE, we compare with \texttt{fippy}, using a Gaussian conditional sampler as the lowest common denominator of available sampling options.
\CRANpkg{vimp} is not included as it only shares the refitting-based methods (LOCO, WVIM) with \CRANpkg{xplainfi}, and differences in supported metrics, learner frameworks, and resampling strategies make a fair apples-to-apples comparison difficult.
The MSE was used as the evaluation metric across all methods, except for \CRANpkg{vip}, which only supports the root mean squared error.
Since RMSE is a monotone transformation of MSE, feature rankings are preserved but scaled importance magnitudes differ.
Both benchmarks are defined in R using \CRANpkg{batchtools} \citep{lang2017batchtoolstools} for cluster-based parallelization, and \CRANpkg{reticulate} is used to run Python implementations.

Each method is evaluated using one of four learners to account for variability in learner capabilities: a linear model (\CRANpkg{stats} / \texttt{scikit-learn} \citep{scikitlearn2011}), a random forest (\CRANpkg{ranger} / \texttt{scikit-learn}, 500 trees), \texttt{XGBoost} \citep{chen2016xgboostscalable} (1000 rounds, early stopping after 50, \(\eta = 0.1\)), and an MLP (\CRANpkg{mlr3torch} / \texttt{scikit-learn}, 1 hidden layer, 20 neurons, 500 epochs with early stopping).
Of these, \texttt{XGBoost} is the only learner with an identical underlying implementation across R and Python.
The linear model is expected to yield approximately identical results in either language.
Learner configurations were chosen for simplicity and robustness rather than predictive performance, i.e., learners were not tuned beyond built-in regularization mechanisms.

\subsection{Importance benchmark}\label{importance-benchmark}

We compare importance scores by scaling values for a given method and DGP to the unit interval to examine relative magnitudes.
A rank-based comparison is available in the online supplement.

We compare five FI methods across one to four different implementations.
Table \ref{tab:tab-benchmark-methods} lists these methods and packages, alongside the parameters used for each method where applicable.
To obtain reliable FI estimates, we used 50 repetitions for PFI and CFI, and similarly a high number of permutations for mSAGE and cSAGE, while enabling the ``early stopping'' options in both \CRANpkg{xplainfi}'s and \texttt{fippy}'s implementations to avoid high computational cost at diminishing returns.
Since we prioritize comparability, we used Gaussian conditional sampling for the conditional methods CFI and cSAGE throughout the experiment, as this is the only sampling mechanism implemented by both \CRANpkg{xplainfi} \emph{and} \texttt{fippy}.
Both packages offer additional sampling mechanisms with greater flexibility, but would have introduced more variability into the comparison.
The tasks selected for this experiment consist of simulation settings with DGPs designed to showcase differences and similarities among feature importance methods; a full listing is provided in the online supplement.
Here, we focus on the ``correlated'' DGP generated by \texttt{xplainfi::sim\_dgp\_correlated()} as introduced in the previous section (\(Y = 2X_1 + X_3 + \varepsilon\), with \(\mathrm{cor}(X_1, X_2) \in \{0.25, 0.75\}\), \(p = 4\)), and the ``bike sharing'' dataset introduced by \citet{fanaee-t2014eventlabeling} (\(p = 12\)), which is frequently used as an example for FI methods \citep[e.g.,][]{covert2020understandingglobalb, blesch2025conditionalfeaturea}.
The bike sharing dataset's categorical features were converted to numeric values, as this was required to ensure comparability between \CRANpkg{xplainfi} and \texttt{fippy} via the Gaussian conditional sampler.
For simulated data, \(n = 5000\) samples were generated and 2/3 of the observations were used for learner training and 1/3 for feature importance calculation as a test set.
The train and test sets were created consistently across the different implementations, i.e., the linear model for \CRANpkg{xplainfi} was trained on the same data as the one for \CRANpkg{vip} or \texttt{fippy} within a replication.
Importance scores shown are aggregated from 25 replications.
This experiment was conducted on a shared Linux server running Ubuntu 24.04 on an AMD EPYC 9554 CPU with 1.41 TiB of RAM.
For additional details on the simulated settings, see the online supplement or \href{https://mlr-org.github.io/xplainfi/articles/simulation-settings.html}{\CRANpkg{xplainfi}'s online documentation}.

\begin{table}
\centering
\caption{\label{tab:tab-benchmark-methods}Methods, implementations, and parameters for both benchmarks. Parameter names correspond to xplainfi's API and are set equivalently in other implementations. For the importance benchmark, early stopping was enabled for mSAGE and cSAGE in xplainfi and fippy; for the runtime benchmark, it was disabled to ensure identical workloads.}
\centering
\begin{tabular}[t]{>{\raggedright\arraybackslash}p{1cm}>{\raggedright\arraybackslash}p{2.5cm}>{\raggedright\arraybackslash}p{9cm}}
\toprule
Method & Packages & Parameters\\
\midrule
PFI & xplainfi, fippy, iml, vip & Importance: n\_repeats = 50\newline Runtime: n\_repeats = \{1, 50\}\\
CFI & xplainfi, fippy & Importance: n\_repeats = 50, sampler: Gaussian\newline Runtime: n\_repeats = \{1, 50\}, sampler: Gaussian\\
mSAGE & xplainfi, fippy & Importance: n\_permutations = 100, n\_samples = 100, min\_permutations = 20\newline Runtime: n\_permutations = \{10, 20, 50\}, n\_samples = \{10, 50, 100\}\\
mSAGE & sage & Importance: n\_samples = 100\newline Runtime: n\_samples = \{10, 50, 100\}\\
cSAGE & xplainfi, fippy & Importance: n\_permutations = 100, n\_samples = 100, min\_permutations = 20, sampler: Gaussian\newline Runtime: n\_permutations = \{10, 20, 50\}, n\_samples = \{10, 50, 100\}, sampler: Gaussian\\
\bottomrule
\end{tabular}
\end{table}

Due to the large number of factors in the experiment, we present a subset of results here and refer to the online supplement for a complete overview.
We focus on the \texttt{linear} and \texttt{boosting} learners since they are the most comparable across R and Python.

\begin{figure}

{\centering \includegraphics[width=0.95\linewidth,alt={Boxplots of scaled importance scores by feature, colored by implementing package. x1 has the highest importance at 100, and x2 and x4 have an importance of 0. vip shows slightly higher importance for x3. xplainfi and fippy show identical values. cSAGE assigns importance to x2 similar in magnitude to x3.}]{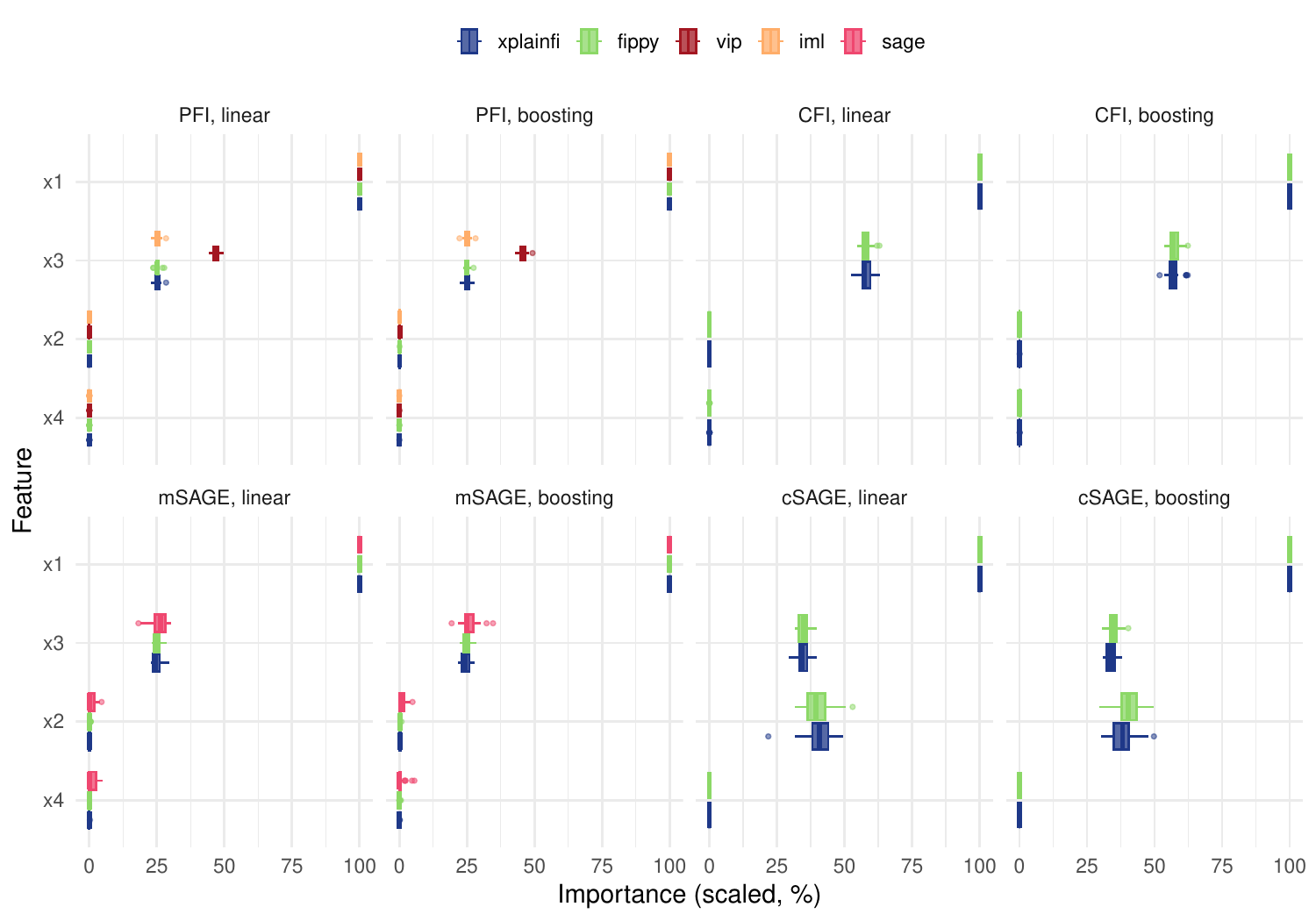} 

}

\caption{Importance scores (scaled to percentages) for PFI, CFI, mSAGE, and cSAGE across implementations on the correlated simulation setting with r = 0.75, based on either the linear model or the boosting learner.}\label{fig:plot-importance-correlated-all}
\end{figure}

Figure \ref{fig:plot-importance-correlated-all} shows almost identical scaled importance scores across all implementing packages for PFI and CFI.
The exception is \CRANpkg{vip}, which produces the same feature ranking but with differing importance scores, most likely due to the different evaluation metric.
The feature ranking is equivalent between the two learner types.
For mSAGE and cSAGE, similarly close agreement is visible between \CRANpkg{xplainfi} and \texttt{fippy}, while \texttt{sage}'s kernel SAGE shows scores close to the other implementations with a slightly larger variance.

\begin{figure}

{\centering \includegraphics[width=0.95\linewidth,alt={Boxplots of scaled importance scores by feature, colored by implementing package. Most features have low or zero importance. `hour` is highest, followed by `year` and `humidity`. `season` ranks lower but is nonzero only for the linear model and PFI. `apparent_temperature` also only scores around 5 with high variance for the linear model. Methods using the boosting learner assign higher importance to `working_day` and `temperature`.}]{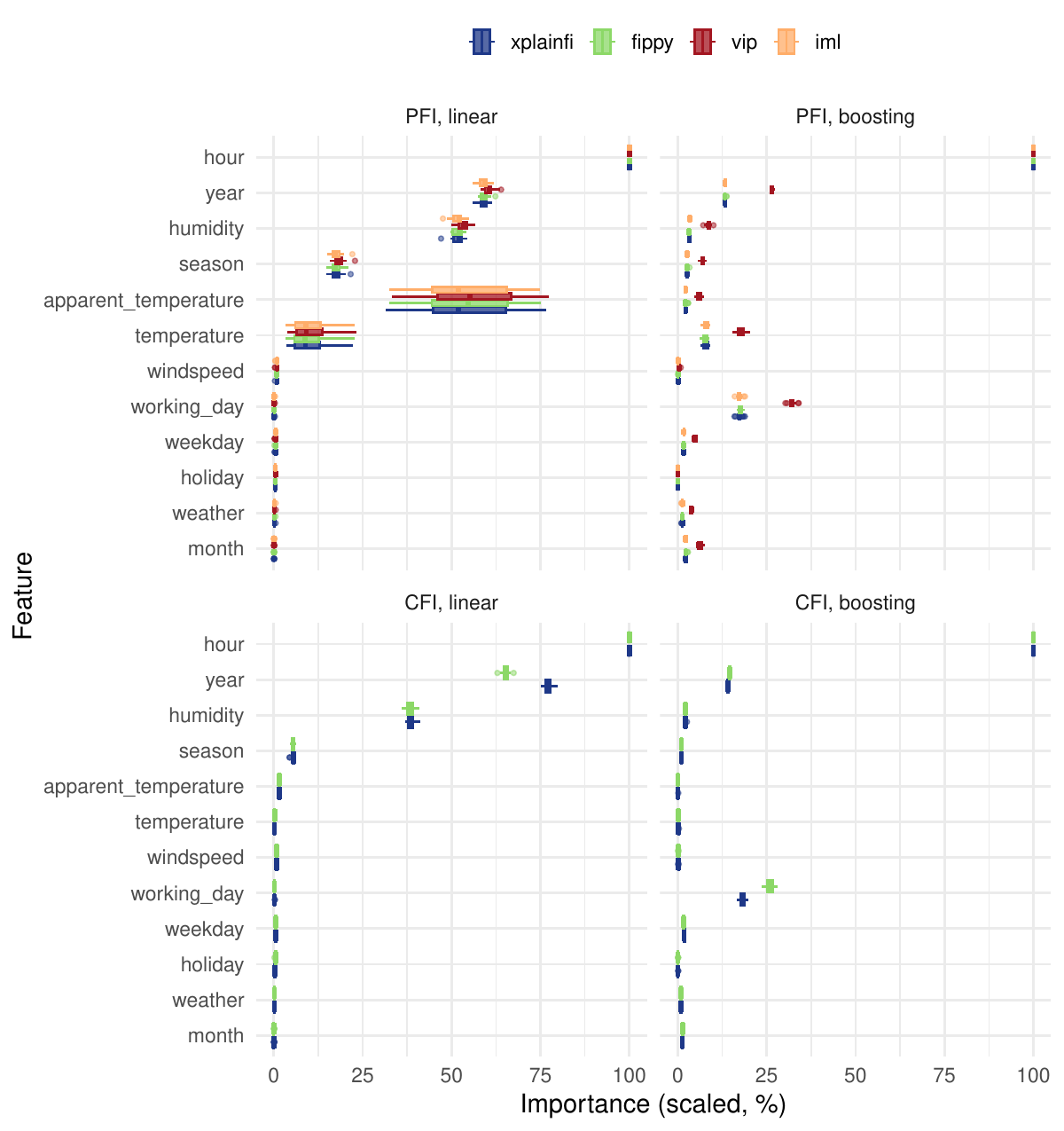} 

}

\caption{Importance scores (scaled to percentages) for PFI and CFI across implementations on the bike sharing dataset, based on either the linear model or the boosting learner.}\label{fig:plot-importance-bikesharing-pficfi}
\end{figure}

\begin{figure}

{\centering \includegraphics[width=0.95\linewidth,alt={Boxplots of scaled importance scores by feature, colored by implementing package. Similar overall picture as in the previous plot, with notably higher importances for temperature and `apparent_temperature` for the methods using the linear model. `sage` (kernel SAGE) disagrees with the other implementations both in ranking and in magnitude, showing `hour` to be close to or below 50, with `apparent_temperature` consistently at 100 when the linear model is used, while hour remains at 100 when the boosting learner is used.}]{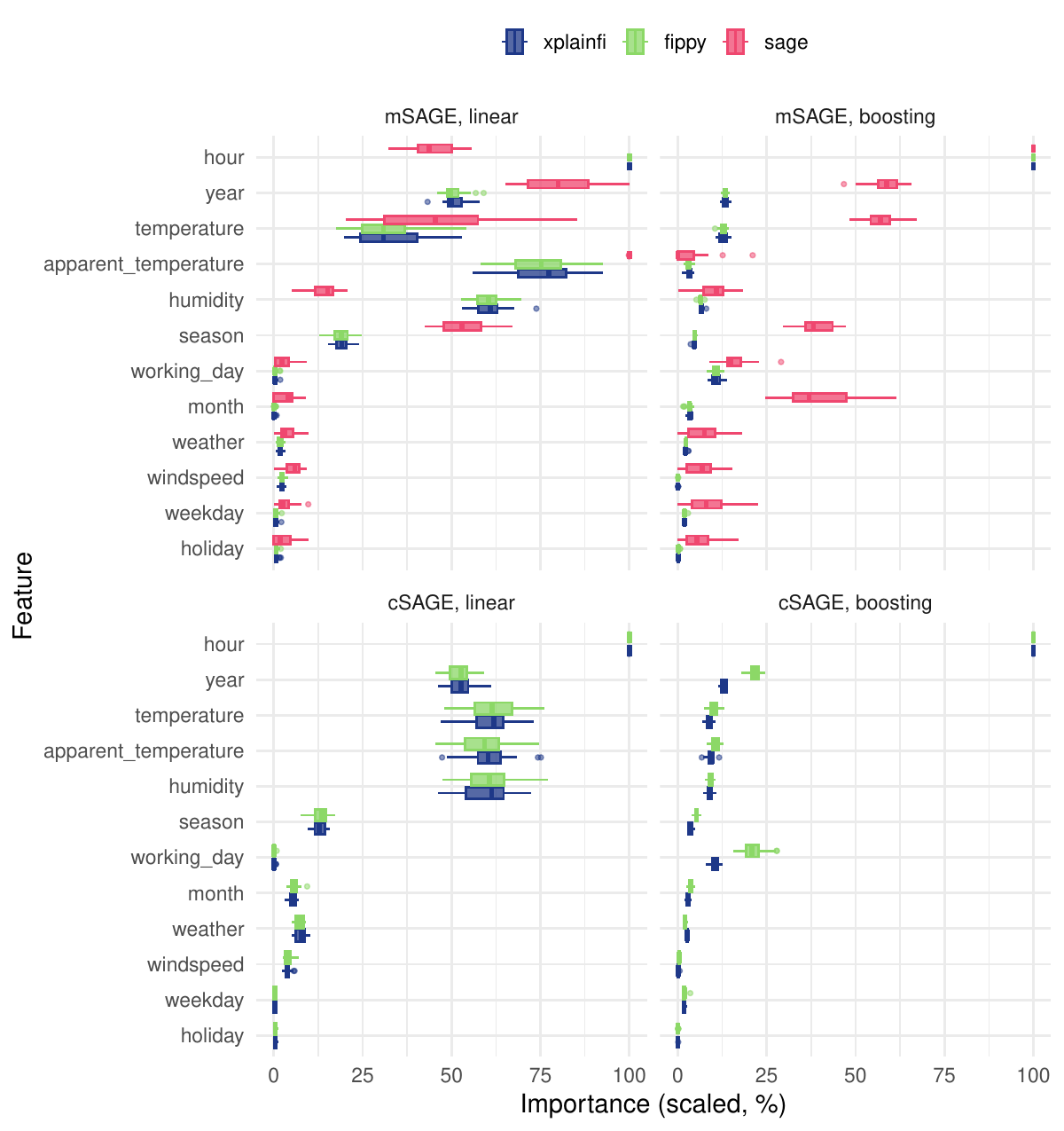} 

}

\caption{Importance scores (scaled to percentages) for mSAGE and cSAGE across implementations on the bike sharing dataset, based on either the linear model or the boosting learner.}\label{fig:plot-importance-bikesharing-sages}
\end{figure}

Next, we consider the bike sharing task.
Figure \ref{fig:plot-importance-bikesharing-pficfi} shows PFI and CFI using the linear model and boosting learner.
For PFI, we see strong agreement between the methods similar to the previous setting.
For CFI, \CRANpkg{xplainfi} and \texttt{fippy} produce different (scaled) importance scores for the \texttt{year} and \texttt{working\_day} features.
This is most likely explained by the Gaussian conditional sampler being implemented slightly differently in both packages, combined with its use here on effectively categorical features encoded as integers, which is not the most appropriate choice.
A similar pattern is visible in Figure \ref{fig:plot-importance-bikesharing-sages} for cSAGE with the same underlying cause but good agreement otherwise.
For mSAGE, \texttt{sage}'s importance scores produce a different pattern which does not appear to agree with the other methods.
Scores for the remaining simulation settings and learners are available in the online supplement.
In all settings, \CRANpkg{xplainfi}'s results agree with those of the reference implementations apart from minor deviations that did not affect the overall ranking, as is also shown by the rank-based analysis in the supplement.

\subsection{Runtime benchmark}\label{runtime-benchmark}

Methods were run exclusively using linear models to isolate the computational cost of the FI method itself from that of the learner (see Table \ref{tab:tab-benchmark-methods} for parameter ranges).
We used only the \texttt{peak} task from \CRANpkg{mlbench}, which provides a regression problem with user-definable sample and feature sizes.
Experiments were run for 25 replications on the Intel Xeon Platinum 8380 compute nodes of the Leibniz Supercomputing Centre.

\begin{figure}

{\centering \includegraphics[width=0.95\linewidth,alt={Pointrange plots of runtime in seconds by method and implementation, colored by implementing package. For PFI and CFI, xplainfi is generally faster than fippy or on par and always faster than iml. For mSAGE, xplainfi is fastest. sage becomes faster the more features are used, and is slowest for 5 features and roughly on par with xplainfi for 20 features.}]{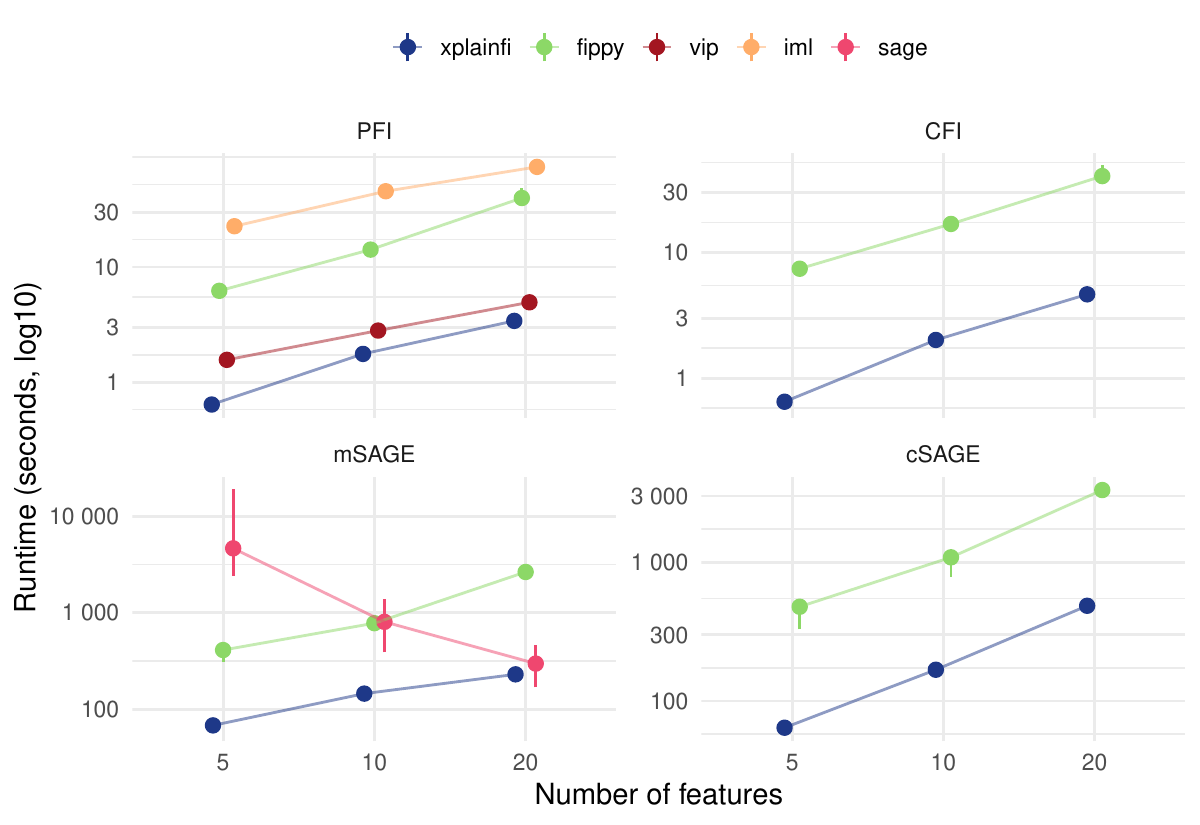} 

}

\caption{Median runtime in seconds with 25\% and 75\% quantiles for PFI and CFI with n\_repeats = 50 and mSAGE and cSAGE with n\_samples = 100 and n\_permutations = 100 across implementations on the peak simulation setting with 5, 10, and 20 features and 5000 samples using a linear model across 25 replications.}\label{fig:plot-runtime-all}
\end{figure}

Figure \ref{fig:plot-runtime-all} shows median runtimes with 25\% and 75\% quantiles for \(n = 5000\) with varying \(p\).
For PFI, \CRANpkg{iml} is clearly the slowest implementation after \texttt{fippy}.
\CRANpkg{vip} shows only slightly slower times than \CRANpkg{xplainfi}.
For mSAGE, \texttt{sage} notably becomes faster with more features, likely because the kernel SAGE implementation converges faster in that case.
Across all methods, \texttt{fippy} is notably slower than \CRANpkg{xplainfi}.
Overall, \CRANpkg{xplainfi} was consistently faster than or comparable to the reference implementations under close-to-equal settings.
Additional results for all methods and parameter configurations are available in the online supplement.

\section{Discussion and conclusion}\label{sec-discussion}

We introduced \CRANpkg{xplainfi}, a package that provides a comprehensive suite of global, loss-based FI methods within a unified framework built on the \CRANpkg{mlr3} ecosystem, by implementing methods previously unavailable or scattered across packages, including conditional feature importance (CFI), relative feature importance (RFI), and both marginal and conditional SAGE.
This allows \CRANpkg{xplainfi} to focus primarily on FI methods without reimplementing common ML building blocks such as model interfaces, resampling, and performance measures.
However, this also means \CRANpkg{xplainfi} is not compatible with related frameworks such as \CRANpkg{tidymodels} or arbitrary ``unwrapped'' models and learner implementations.
We accept this compromise because we believe we would not be able to maintain or expand the current set of features otherwise.
Our benchmark experiments demonstrate that \CRANpkg{xplainfi} produces importance scores consistent with reference implementations across multiple DGPs and learner types, while offering competitive runtime performance.

Choosing the ``best'' methods among those available depends on the research question and the type of association of interest (marginal vs.~conditional), and in no small part on the computational budget available, which can affect the choice between perturbation-based, refit-based, and Shapley-based methods.
For detailed guidance on underlying estimands and interpretation, we refer to \citet{ewald2024guidefeatureb}.

The modular feature sampler architecture is unique, enabling \CRANpkg{xplainfi} to perform conditional FI analysis on continuous and mixed data alike via flexible samplers.
This includes the ability to specify arbitrary conditioning sets for each sampler in the \texttt{Conditional} family, which enables conditional SAGE and RFI.
Interest in conditional methods has been growing, and future work may benefit from the building blocks available with \CRANpkg{xplainfi}.
However, these methods come with trade-offs: the ARF sampler, while flexible, incurs higher computational cost and memory usage than Gaussian sampling, and SAGE methods can be slow to compute due to the large number of feature coalitions that must be evaluated.
For the latter, we aim to implement corresponding improvements in the near future.

The package could and should be extended to include additional confidence interval methods, or more generally, uncertainty quantification methods for FI.
Although this is very desirable from a practical perspective, not enough (established) techniques and coverage studies exist, and we think exposing users to unvalidated techniques is not appropriate; such dedicated studies are out of scope for the current paper.

The current package does not offer any visualization options, as we have focused more on the computational aspects.
Often, concrete visualizations for reports and publications have slightly different requirements and are customized accordingly.
Because we provide well-structured container types for our results, the generated FI values (and other results) can be easily plotted using, e.g., \CRANpkg{ggplot2}.

\section{Acknowledgements}\label{acknowledgements}

\setlength{\parindent}{0pt}

The authors gratefully acknowledge the computational and data resources provided by the Leibniz Supercomputing Centre (www.lrz.de).

Lukas Burk is supported by the Federal Ministry of Research, Technology and Space (BMFTR), grant number 01EQ2409E.

Marvin N. Wright is supported by the German Research Foundation (DFG), grant numbers 437611051 and 459360854, and the Federal Ministry of Research, Technology and Space (BMFTR), grant number 01EQ2409E.

Bernd Bischl is supported by the German Research Foundation (DFG), grant number 460135501.

\bibliography{RJreferences.bib}

\address{%
Lukas Burk\\
1) Leibniz Institute for Prevention Research and Epidemiology - BIPS 2) LMU Munich, Department of Statistics 3) Munich Center for Machine Learning (MCML) 4) University of Bremen\\%
Achterstraße 30, 28359 Bremen, Germany\\
\url{https://lukasburk.de}\\%
\textit{ORCiD: \href{https://orcid.org/0000-0001-7528-3795}{0000-0001-7528-3795}}\\%
\href{mailto:burk@leibniz-bips.de}{\nolinkurl{burk@leibniz-bips.de}}%
}

\address{%
Fiona Katharina Ewald\\
1) LMU Munich, Department of Statistics 2) Munich Center for Machine Learning (MCML)\\%
Department of Statistics\\ Chair of Statistical Learning and Data Science\\
\textit{ORCiD: \href{https://orcid.org/0009-0002-6372-3401}{0009-0002-6372-3401}}\\%
\href{mailto:fiona.ewald@lmu.de}{\nolinkurl{fiona.ewald@lmu.de}}%
}

\address{%
Giuseppe Casalicchio\\
1) LMU Munich, Department of Statistics 2) Munich Center for Machine Learning (MCML)\\%
Department of Statistics\\ Chair of Statistical Learning and Data Science\\
\textit{ORCiD: \href{https://orcid.org/0000-0001-5324-5966}{0000-0001-5324-5966}}\\%
}

\address{%
Marvin N. Wright\\
1) Leibniz Institute for Prevention Research and Epidemiology - BIPS 2) University of Bremen\\%
Achterstraße 30, 28359 Bremen, Germany\\
\textit{ORCiD: \href{https://orcid.org/0000-0002-8542-6291}{0000-0002-8542-6291}}\\%
}

\address{%
Bernd Bischl\\
1) LMU Munich, Department of Statistics 2) Munich Center for Machine Learning (MCML)\\%
Department of Statistics\\ Chair of Statistical Learning and Data Science\\
\textit{ORCiD: \href{https://orcid.org/0000-0001-6002-6980}{0000-0001-6002-6980}}\\%
}

\end{article}

\end{document}